\documentclass[12pt]{article}
\usepackage[utf8]{inputenc}
\usepackage{amsmath}
\usepackage{amssymb}
\usepackage{amsfonts}
\usepackage{bm}
\usepackage{bbm}
\usepackage{placeins}
\usepackage[hidelinks]{hyperref}
\usepackage{booktabs}
\usepackage[left=1in, top=1in, right=1in, bottom=1in, nohead, paperwidth=8.5in, paperheight=11in]{geometry}
\usepackage[noend]{algpseudocode}
\usepackage{algorithm}
\usepackage{natbib}
\usepackage{graphicx}
\usepackage{xcolor}

\usepackage{multirow}
\usepackage{tabularx}
\usepackage{siunitx}
\usepackage{etoolbox}
\newrobustcmd\B{\DeclareFontSeriesDefault[rm]{bf}{b}\bfseries}

\usepackage[skip=0pt,format=hang]{caption}
\newcommand{\ma}[1]{\ensuremath{\mathbf{#1}}}
\newcommand{\indicator}[1]{\ensuremath{I}[ #1 ]}
\newcommand{\fracpartial}[2]{\frac{\partial #1}{\partial #2}}
\newcommand{\dep}[2]{\delta(\ma{c}_{#1 #2}, p_{i #2})}

\newcommand{\yhat}{\hat{y}}

\allowdisplaybreaks

\title{Classifier Chain Networks for Multi-Label Classification}
\author{Daniel J.W. Touw, Michel van de Velden}
\date{}

\begin{document}

\maketitle

\begin{abstract}
    \noindent
    The classifier chain is a widely used method for analyzing multi-labeled data sets. In this study, we introduce a generalization of the classifier chain: the \emph{classifier chain network}. The classifier chain network enables joint estimation of model parameters, and allows to account for the influence of earlier label predictions on subsequent classifiers in the chain. Through simulations, we evaluate the classifier chain network's performance against multiple benchmark methods, demonstrating competitive results even in scenarios that deviate from its modeling assumptions. Furthermore, we propose a new measure for detecting conditional dependencies between labels and illustrate the classifier chain network's effectiveness using an empirical data set.

    \bigskip

\end{abstract}

\section{Introduction}
A multi-label classifier models the association of an observation with multiple labels. In contrast to binary and multi-class classification, where each observation in the data is assigned to a single class, an observation in a multi-label classification task can have multiple labels. This type of problem arises in different fields, such as object detection in images, text analysis, bioinformatics, and recommendation systems \citep{tsoumakas2010mining}. Consequently, numerous methods have been developed to handle multi-labeled outcomes. In contrast to existing methods, which often focus on modeling each outcome variable separately, our proposed method jointly models all labels to capture dependencies between them. In this study, we also refer to these dependencies between labels as \emph{label interdependencies}.

A frequently used method for a classification task with multi-labeled outcomes is to decompose the task into separate independent binary classifications \citep[e.g.,][]{boutell2004learning, luaces2012binary}. This approach is typically referred to as \emph{binary relevance}. A limitation of binary relevance is the fact that it does not exploit potential correlations between the different labels \citep{godbole2004discriminative, zhang2013review}. Empirical evidence supports the notion that models explicitly accounting for possible dependencies between labels may outperform binary relevance \citep[e.g.,][]{ji2008sharedsubspace, lozamencia2008efficient, liu2015comparisons}. A number of methods have been developed that simultaneously model a multi-labeled outcome and take into account label interdependencies. Early examples are AdaBoost.MH \citep{schapire1998adaboost}, multi-label decision trees \citep{clare2001ml-dt}, multi-label $k$-nearest neighbors \citep{zhang2005ml-knn}, random $k$-labelsets \citep{tsoumakas2011random}, instance-based learning combined by logistic regression \citep{cheng2009combining}, and classifier chains \citep{read2009ecc, read2011ecc}. In this study, we focus on the latter because it has shown favorable results in the existing literature \citep[e.g.,][]{moyano2018review, rivolli2020empirical}.

The classifier chain is closely related to binary relevance, as both methods use binary classifiers to predict the occurrence of each of the $L$ labels. However, it extends binary relevance by accommodating label interdependencies \citep{zhang2018overview}. This is achieved by assuming a sequential order in the occurrence of the labels and chaining the classifiers accordingly, each using the preceding labels as additional explanatory variables. As a result, each label is modeled using more information than its predecessors. A notable property of the classifier chain is the reliance on the ordering of the labels, which may not always be known in practical applications. \cite{read2009ecc, read2011ecc} address this challenge by introducing an ensemble of classifier chains, combining multiple chains, each with randomly ordered sequences of labels.

Classifier chains have shown to be a strong performer for multi-label classification tasks. For example, \citet{moyano2018review} and \citet{bogatinovski2022comprehensive} demonstrated that the ensemble classifier chain is among the top performers across various evaluation metrics. This observation has led to a number of studies seeking to expand and improve the method. Variations are, among others, probabilistic classifier chains \citep{dembczynski2010bayes}, recurrent classifier chains \citep{nam2017maximizing}, and nested stacking \citep{senge2019rectifying}. Furthermore, \cite{weng2020label} show that label-specific features can improve out-of-sample performance and \cite{jun2019conditional} propose a method to forgo ensembles of classifier chains and order the labels based on their conditional entropy.

In this paper, we propose a generalization of the classifier chain that conceptualizes the model as a single network. Unlike traditional classifier chains, which fit classifiers sequentially and are ``blind'' to the impact of their predictions on subsequent classifiers, our approach models all outcomes simultaneously. Consequently, our model is able to take into account the impact of the prediction for the first label on the prediction for the last label. We call this generalization the \emph{classifier chain network}. Our approach aligns with previous studies, such as \cite{nam2017maximizing}, which treat multi-label classification tasks as viable targets for neural network architectures. However, in contrast to deep learning methodologies, we use modeling assumptions to constrain the number of parameters, ensuring an interpretable model applicable to small to medium-sized data sets. As part of the proposed generalization, we incorporate a regularization step into the loss function used to estimate the model parameters. This adjustment, inspired by \cite{burg2016gensvm}, aims to mitigate an overemphasis on observations with many misclassified labels.

To gain insights into the performance of the classifier chain network relative to other multi-label classification methods, we conduct a comprehensive simulation study. These simulations provide a controlled environment where the ground truth is known, allowing us to identify specific scenarios and conditions that highlight the strengths and limitations of the classifiers under consideration. The results show that the classifier chain network generally outperforms benchmark methods across multiple performance metrics and simulation designs. Notably, it achieves excellent results in terms of the negative log-likelihood, reflecting a desirable balance of higher confidence in correct predictions and appropriately lower confidence in incorrect ones.

Exploring an issue related to multi-label classification tasks, we examine in which scenarios methods that explicitly account for label interdependencies offer benefits over binary relevance. Prior research has noted cases where binary relevance demonstrates competitive performance \citep[e.g.][]{luaces2012binary, douibi2019Analysis}. Additionally, it has been established that for methods explicitly modeling label interdependencies to be advantageous, there must be conditional dependency between the labels \citep{dembczynski2010regret}. We propose a new measure for detecting conditional dependencies and compare it to existing approaches.

The remainder of this paper is structured as follows. In Section~\ref{sec:ccn}, we present the classifier chain network and its estimation procedure. To compare our method with alternative multi-label classification methods, a simulation study is performed in Section~\ref{sec:simulations} and we propose a new method to detect conditional dependencies between labels in Section~\ref{sec:conditionaldeps}. In Section~\ref{sec:application}, we present an application of the classifier chain network to empirical data, and Section~\ref{sec:conclusion} concludes the paper.

\section{Classifier Chain Network}
\label{sec:ccn}
This section provides an overview of our proposed method, the classifier chain network. In contrast to traditional classifier chains that fit classifiers sequentially and do not consider that each prediction influences subsequent labels, the classifier chain network models all label outcomes simultaneously. In Section~\ref{sec:ccn_model}, we outline the general framework for the simultaneous modeling of multi-labeled outcomes. Following that, Section~\ref{sec:ccn_estimation} details the parameter estimation procedure for the classifier chain network, highlighting the flexible loss function that accommodates various modeling choices.

\subsection{Model}
\label{sec:ccn_model}
We use the following notation. Let the $n \times m$ matrix \ma{X} with rows $\ma{x}_i^\top$ contain the explanatory variables for each observation $i$, and let the $n \times L$ binary matrix \ma{Y} contain the true labels. A label is referred to as positive (negative) if its value is one (zero). Following standard classifier chain conventions, each label can affect the prediction of consecutive labels, but not those preceding it. The classifier chain network follows the approach presented in \cite{dembczynski2010bayes}, who propagate the continuous predictions through the chain of classifiers instead of the binary outcomes as done by \cite{read2009ecc, read2011ecc}.

Using an activation function $\alpha(\cdot)$, a continuous prediction for the $\ell$\textsuperscript{th} label of the $i$\textsuperscript{th} observation is computed.
For the first label, no additional information originating from other labels is available. Hence, we model the continuous prediction for this label as
\begin{equation}
\label{eq:label1}
    p_{i1} = \alpha(b_1 + \ma{x}_i^\top \ma{w}_1),
\end{equation}
where $b_1$ is a bias term and $\ma{w}_1$ is a vector of weights. Note that, for ease of exposition, we assume that the input of the activation function is affine in $\ma{x}_i$. In practice, the matrix \ma{X} can be extended with nonlinear functions of the original variables to allow for certain nonlinear effects.

\begin{figure}
\includegraphics{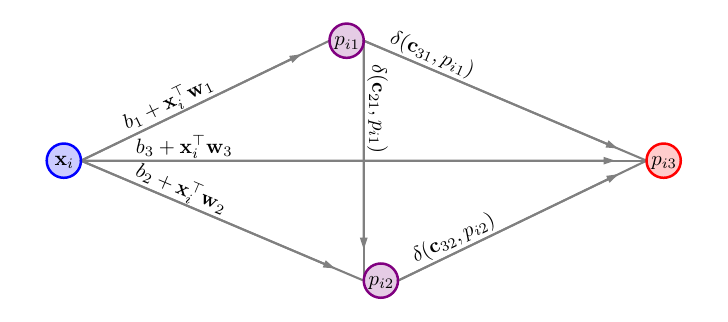}
\centering
\caption{A visual representation of a classifier chain network with three labels. The blue node provides the input vector $\mathbf{x}_i$ that undergoes an affine transformation on each of its outgoing edges. At the purple and red nodes, the values of the incoming edges are summed and used in an activation function to produce a prediction $p_{i\ell}$ for the corresponding labels. The purple nodes fulfil a dual role, as they compute part of the output of the network and provide (a transformation of) this output to subsequent nodes.}
\label{fig:ccn}
\end{figure}

To model the label interdependencies, $d$-dimensional parameter vectors $\ma{c}_{k\ell}$, where $k>\ell$, are introduced. These parameters are used to capture the effect of label $\ell$ on label $k$, with $\dep{k}{\ell}$ representing this relationship.
The predictions for the remaining $L-1$ labels are then modeled by
\begin{align}
    p_{i2} &= \alpha(b_2 + \ma{x}_i^\top \ma{w}_2 + \dep{2}{1}) \nonumber\\
    p_{i3} &= \alpha(b_3 + \ma{x}_i^\top \ma{w}_3 + \dep{3}{1} + \dep{3}{2}) \nonumber\\
    &\;\; \vdots \nonumber\\
    p_{iL} &= \alpha\left(b_L + \ma{x}_i^\top \ma{w}_L + \sum_{\ell=1}^{L-1} \dep{L}{\ell}\right). \label{eq:labelL}
\end{align}
To arrive at binary predictions, an appropriate thresholding function must be used on the predictions $p_{i\ell}$. The resulting network structure is visualized for a scenario with three labels in Figure~\ref{fig:ccn}, where the colors of the nodes represent their respective (dual) roles. In the next section, we consider the estimation of the model parameters.

\subsection{Estimation}
\label{sec:ccn_estimation}
A key aspect of the proposed classifier chain network is the joint estimation of all model parameters. To facilitate this estimation, we need to define a loss function. Since there is no single method for doing this, we introduce general notation in Section~\ref{sec:ccn_loss} and present multiple options in Section~\ref{sec:ccn_choices}. The optimization procedure, along with a discussion of the tuning parameters, is provided in Section~\ref{sec:opt_tuning}.

\subsubsection{Loss}
\label{sec:ccn_loss}
To evaluate each individual prediction $p_{i\ell}$, a per-label loss $h(p_{i\ell})$ can be used. One approach to aggregate these values is to sum them over all observations $i$ and labels $\ell$. However, this method may overemphasize observations with many misclassified labels. To address this issue and retain flexibility, we follow the approach of \cite{burg2016gensvm}, who faced a similar problem in the context of multiclass support vector machines. Specifically, we use the $\ell_q$-norm of the losses for a single observation $i$,
\begin{equation*}
    L_i(\ma{b}, \ma{W}, \ma{C}) = \left(
        \sum_{l=1}^L |h(p_{i\ell})|^q
    \right)^{1/q},
\end{equation*}
with $q \geq 1$. Larger values for $q$ assign more weight to labels for which the misclassification error as defined by $h(p_{i\ell})$ is larger. In the limit, the loss for observation $i$ is $\lim_{q\rightarrow\infty} L_i(\ma{b}, \ma{W}, \ma{C}) = \max_\ell h(p_{i\ell})$. Note that using the $\ell_q$-norm in this context requires $h(p_{i\ell}) \geq 0$. Combining the loss for all $n$ observations, the loss function of the classifier chain network can be formulated as
\begin{equation}
\label{eq:ccn_loss}
    L(\ma{b}, \ma{W}, \ma{C}) = \frac{1}{nL^{1 / q}} \sum_{i=1}^n 
    \left(
        \sum_{\ell=1}^L |h(p_{i\ell})|^q
    \right)^{1/q} + \frac{\lambda}{r} \left( \| \ma{W} \|^2 + \| \text{vec}(\ma{C}) \|^2 \right),
\end{equation}
where the $L \times m$ matrix \ma{W} has rows $\ma{w}_i^\top$, \ma{C} is a three-dimensional $L \times L \times d$ array which holds the $d$-dimensional vectors $\ma{c}_{k\ell}$, $\lambda / r \left( \| \ma{W} \|^2 + \| \text{vec}(\ma{C}) \|^2 \right)$ is a penalty term to avoid overfitting, $\lambda \geq 0$ is the regularization parameter, and $r$ is the number of elements in $\ma{W}$ and $\ma{C}$. In our notation, $\| \cdot \|$ denotes either the $\ell_2$-norm of a vector or the Frobenius norm of a matrix and $\text{vec}(\cdot)$ is the vectorization operation.

\subsubsection{Modeling Choices}
\label{sec:ccn_choices}
Section~\ref{sec:ccn_model} covered the general form of the classifier chain network. An important detail is the dependency structure of the labels, as defined by $\dep{k}{\ell}$. A straightforward choice, which is used in the remainder of this paper, is a scalar multiplication of the form
\begin{equation}
\label{eq:ccn_dependency}
    \dep{k}{\ell} = c_{k\ell}p_{i\ell},
\end{equation}
which imposes a constant effect of $p_{i\ell}$ on $p_{ik}$ that is either positive or negative, depending on the sign of $c_{k\ell}$. However, a more flexible model can be obtained by using, for example, a polynomial of degree two or higher. Note that binary relevance, which is a collection of $L$ independent binary classifiers, is a special case of the classifier chain network where $\dep{k}{\ell}=0$ for all $k$ and $\ell$.

Other modeling choices include the activation function $\alpha(\cdot)$ and the quantification of the per-label loss $h(p_{i\ell})$. The remainder of this section outlines two general approaches for these elements, illustrated with specific examples.

\textbf{Probabilistic.}
The first approach involves methods that produce predictions in the range $(0,1)$, which are often interpreted as probabilities. A well-known example of this is the sigmoid activation function
\begin{equation}
\label{eq:sigmoid}
    p_{i\ell} = \alpha(\theta_{i\ell}) = \frac{1}{1+\exp(-\theta_{i\ell})},
\end{equation}
where the input is denoted by $\theta_{i\ell}$. A loss function that can be applied to predicted values in this range is
\begin{equation*}
    h(p_{i\ell}) =
    \begin{cases}
        (1-p_{i\ell})^{\xi^+} \log(p_{i\ell}) & \text{if $y_{i\ell} = 1$} \\
        p_{i\ell}^{\xi^-} \log(1 - p_{i\ell}) & \text{if $y_{i\ell} = 0$}
    \end{cases}
\end{equation*}
with $\xi^+, \xi^- \geq 0$ and where setting $\xi^+ = \xi^- = 0$ corresponds to binary cross-entropy, $\xi^+ = \xi^- \neq 0$ to focal loss \citep{lin2020focal}, and $\xi^+ \neq \xi^-$ to asymmetric focusing \citep{ridnik2021asymmetric}. 

\textbf{Margin-based.} 
The second approach is based on support vector machines \citep{cortes1995support}, where a separating hyperplane is used to classify observations. The corresponding activation function is
\begin{equation*}
    p_{i\ell} = \alpha(\theta_{i\ell}) = \theta_{i\ell},
\end{equation*}
which can be converted into a binary prediction by $\hat{y}_{i\ell} = \indicator{p_{i\ell} \geq 0}$. Examples of corresponding loss functions are the absolute, quadratic, and Huber hinge losses \citep[e.g.,][]{rosset2007piecewise, groenen2008svm}. For example, the Huber hinge loss can be defined as
\begin{equation*}
    h(p_{i\ell}) =
    \begin{cases}
        1-(2y_{i\ell}-1)p_{i\ell}-\frac{\kappa+1}{2} & \text{if $(2y_{i\ell}-1)p_{i\ell}\leq \kappa$} \\
        \frac{1}{2(\kappa+1)}\max(0, 1-(2y_{i\ell}-1)p_{i\ell})^2 & \text{otherwise},
    \end{cases}
\end{equation*}
with tuning parameter $\kappa>-1$. If $\kappa \downarrow -1$, this function approaches the absolute hinge loss.

\subsubsection{Optimization and Tuning}
\label{sec:opt_tuning}
To minimize the loss function in \eqref{eq:ccn_loss}, we use an iterative algorithm based on the Broyden-Fletcher-Goldfarb-Shanno (BFGS) method. This quasi-Newton method has favorable convergence properties, even for nonconvex objective functions \citep{dennis1977quasi}. Moreover, the complexity of each iteration is quadratic in the number of parameters, making it more suitable than Newton's method for larger data sets. Below, we present a general description of the optimization procedure, for a more detailed explanation, see, for example, Chapter~6.1 of \citet{nocedal1999numerical}.

The steps taken in iteration $t$ of the algorithm (presented in Algorithm~\ref{alg:ccn_BFGS}) are as follows. First, a descent direction is computed using an approximation of the inverse of the Hessian and the gradient of the loss function, denoted by $\ma{H}^{(t)}$ and $\nabla L(\ma{b}^{(t)}, \ma{W}^{(t)}, \ma{C}^{(t)})$, respectively. The derivation of the gradient is provided in Appendix~\ref{appendix:gradients}. Second, a line search is performed to find a step size $s^*$ that satisfies the Wolfe conditions \citep{wolfe1969convergence, wolfe1971convergence}, which is then used in combination with the descent direction to update the parameters. The Wolfe conditions require two constants corresponding to the Armijo rule \citep{armijo1966minimization} and curvature condition. For the computations in this paper, these are set to $c_1=10^{-6}$ (Armijo rule) and $c_2=0.9$ (curvature condition). Third, the approximation of the Hessian is updated using the updated values of the parameters and gradient (for the approximation at $t=1$, the identity matrix $\ma{I}$ can be used). These three steps are repeated until convergence, determined by the tolerance $\varepsilon_c$, which is set to $10^{-6}$.

\begin{algorithm}
\caption{Pseudo-code for the BFGS algorithm that minimizes the loss function for the classifier chain network.\label{alg:ccn_BFGS}}
\hspace*{\algorithmicindent}\:\textbf{Input:}\:Initial estimates for $\ma{b}^{(1)}$, $\ma{W}^{(1)}$, and $\ma{C}^{(1)}$, Wolfe condition constants $c_1$ and $c_2$, 
\hspace*{\algorithmicindent}\:\phantom{\bf Input:}\:data matrices $\ma{X}$ and $\ma{Y}$, tuning parameters $q$ and $\lambda$, convergence threshold $\varepsilon_c$\\
\hspace*{\algorithmicindent}\:\textbf{Output:}\:$\widehat{\ma{b}}$, $\widehat{\ma{W}}$, and $\widehat{\ma{C}}$ that minimize $L(\ma{b}, \ma{W}, \ma{C})$
\begin{algorithmic}[1]
\State $\ma{H}^{(1)} \leftarrow \ma{I}$
\State $t \leftarrow 1$
\While{$L(\ma{b}^{(t-1)}, \ma{W}^{(t-1)}, \ma{C}^{(t-1)}) / L(\ma{b}^{(t)}, \ma{W}^{(t)}, \ma{C}^{(t)}) - 1 > \varepsilon_c$}
    \State $\ma{d}^{(t)} \leftarrow - \ma{H}^{(t)} \nabla L(\ma{b}^{(t)}, \ma{W}^{(t)}, \ma{C}^{(t)})$
    \State Find step size $s^*$ that satisfies the Wolfe conditions
    \State $\bigl[\ma{b}^{(t+1)}, \ma{W}^{(t+1)}, \ma{C}^{(t+1)}\bigr] \leftarrow \bigl[\ma{b}^{(t)}, \ma{W}^{(t)}, \ma{C}^{(t)}\bigr] + s^* \ma{d}^{(t)}$
    \State Compute the update $\ma{H}^{(t+1)}$
    \State $t \leftarrow t + 1$
\EndWhile
\State $\bigl[\widehat{\ma{b}}, \widehat{\ma{W}}, \widehat{\ma{C}} \bigr] \leftarrow \bigl[ \ma{b}^{(t)}, \ma{W}^{(t)}, \ma{C}^{(t)} \bigr]$
\end{algorithmic}
\end{algorithm}

The remaining inputs to Algorithm~\ref{alg:ccn_BFGS} are $q$, which dictates how the loss values for the individual label predictions are aggregated, and $\lambda$, which controls the strength of the penalty term that avoids overfitting. These parameters can be determined using cross-validation in combination with an appropriate out-of-sample scoring metric. Examples of metrics that can be used in the context of multi-label classification are the Hamming and zero-one losses
\begin{align*}
    L_\text{H} &= \frac{1}{nL} \sum_{i=1}^n \sum_{\ell=1}^L \indicator{\yhat_{i\ell} \neq y_{i\ell}}, \\
    L_{0/1} &= \frac{1}{n} \sum_{i=1}^n \indicator{\hat{\mathbf{y}}_i \neq \mathbf{y}_i}
\end{align*}
and the macro-F1 and micro-F1 scores
\begin{align*}
    \text{F1}_\text{macro} &= \frac{1}{L} \sum_{\ell=1}^L \frac{2\text{TP}_\ell}{2\text{TP}_\ell + \text{FP}_\ell + \text{FN}_\ell}\\
    \text{F1}_\text{micro} &= \frac{2\text{TP}}{2\text{TP} + \text{FP} + \text{FN}},
\end{align*}
which are based on the true positives (TP), false positives (FP), and false negatives (FN). The macro-F1 score averages the standard F1 score over the labels, whereas the micro-F1 score does not distinguish between labels.
For probabilistic predictions, one can consider the (average) negative log-likelihood
\begin{equation*}
    L_\text{nll} = -\frac{1}{nL} \sum_{i=1}^n \sum_{\ell=1}^L y_{i\ell} \log(p_{i\ell}) + (1-y_{i\ell}) \log(1 - p_{i\ell}).
\end{equation*}
Finally, it should be noted that the nonconvexity of the loss function in \eqref{eq:ccn_loss} introduces variability in the parameter estimates. To reduce this variability, multiple random starts can be used and the solution with the lowest value for the loss function can be selected. Additionally, we propose an informed initialization approach for parameter estimations, which can enhance the consistency of the algorithm. Specifically, if the classifier chain with the same parameterization as the classifier chain network involves a sequence of models with convex loss functions, the parameter estimates from this classifier chain can serve as a reproducible starting point for the classifier chain network. Combining the informed initialization with random starts can provide a more robust approach to reducing variability and improving overall performance.

\section{Experiments}
\label{sec:simulations}
We conduct a simulation study to investigate the performance of the classifier chain network under different conditions. Our study includes basic simulation designs that align with the modeling assumptions of the classifier chain network, as well as scenarios where the label order used to estimate the model parameters is incorrect or label outcomes are determined under different assumptions. Note that, we focus on designs with small to moderate amounts of data relative to the number of parameters. The simulation designs are detailed in Section~\ref{subsec:sim_designs}. We compare the classifier chain network's performance with that of other methods for which software implementations are publicly available (in Section~\ref{subsec:sim_methods} we briefly review these methods). The results of our simulations are presented in Section~\ref{subsec:sim_results}.

\subsection{Simulation Designs}
\label{subsec:sim_designs}
We consider several simulation designs to examine how the following factors influence the analysis of multi-labeled data: (i) the strength of label interdependencies, (ii) the number of labels, (iii) the correct or incorrect specification of the label order, and (iv) interdependencies based on actual labels versus label probabilities. By varying these factors, we create five distinct data generating processes that comprehensively test the performance of the multi-label classification methods under consideration. In the remainder of this section, we provide a detailed description of these designs.

In all simulations, we consider data sets with $n=200$ observations for $m=3$ explanatory variables and a varying number of labels $L$. The simulated data for the explanatory variables, denoted by the $200\times3$ matrix \ma{X}, are generated using a multivariate normal distribution with a mean of zero and covariance matrix
\begin{equation*}
    \ma{\Sigma} = \left[\begin{array}{@{}rrr@{}}
        2.0 & 0.4 & 0.4 \\
        0.4 & 2.0 & 0.4 \\
        0.4 & 0.4 & 2.0
    \end{array}\right].
\end{equation*}
The label probabilities are computed using equations \eqref{eq:label1} and \eqref{eq:labelL}, with the sigmoid function in \eqref{eq:sigmoid} serving as the activation function $\alpha(\cdot)$. The dependencies between the labels are defined by \eqref{eq:ccn_dependency}. To obtain binary outcomes, denoted by the $200\times L$ label matrix \ma{Y}, the probabilities are transformed using the Bernoulli distribution. It is important to note that, unless otherwise specified, all label probabilities are computed prior to the transformation, thus allowing for interdependencies based on probabilities. This procedure contrasts with the implicit assumption of the classifier chain \citep{read2009ecc, read2011ecc}, where label probabilities are first dichotomized before they influence subsequent label probabilities.

\begin{figure}
\includegraphics{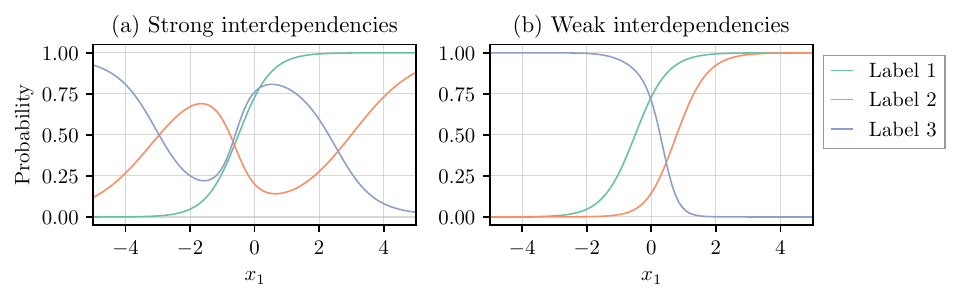}
\centering
\caption{The label probabilities versus the value of $x_1$ under the data generating process containing strong label interdependencies (left), defined by the parameters in \eqref{eq:dgp_strong}; and weak interdependencies (right), defined by the parameters in \eqref{eq:dgp_weak}.}
\label{fig:weak_strong}
\end{figure}

{\bf Baseline designs.}
We use the baseline designs to highlight scenarios where binary relevance performs comparably to methods that explicitly account for label interdependencies, as noted by \citet{luaces2012binary}. To achieve this, we create two variants of the same data generating process. One exhibits strong label interdependencies, where explicitly modeling these interdependencies is beneficial and binary relevance is expected to underperform. The other scenario involves weak interdependencies, where methods that explicitly model interdependencies are not expected to provide added value. The parameters to generate data with strong label interdependencies are
\begin{equation}
\label{eq:dgp_strong}
    \ma{b}^{(\text{s})} = \left[\begin{array}{@{}r@{}}
        1.0 \\
        3.0 \\
        0.5
    \end{array}\right]
    \quad
    \ma{W}^{(\text{s})} = \left[\begin{array}{@{}rrr@{}}
        2.0 & 0.0 & 0.0 \\
        1.0 & 0.0 & 0.0 \\
        -0.5 & 0.0 & 0.0
    \end{array}\right]
    \quad
    \ma{C}^{(\text{s})} = \left[\begin{array}{@{}rrr@{}}
        \text{-}\phantom{0} & \text{-}\phantom{0} & \text{-}\phantom{0} \\
        -6.0 & \text{-}\phantom{0} & \text{-}\phantom{0} \\
        2.0 & -4.0 & \text{-}\phantom{0}
    \end{array}\right],
\end{equation}
where the element $w_{\ell j}^{(s)}$ of $\ma{W}^{(s)}$ is the effect of the $j$\textsuperscript{th} variable in \ma{X} on label $\ell$, and the matrix $\ma{C}^{(\text{s})}$ contains the parameters that determine the direct effect of one label on another. For example, the element $c^{(s)}_{21}$ represents the (constant) effect of the first label on the second. Note the absence of elements on the diagonal and in the upper triangular part of $\ma{C}^{(s)}$ as there is no effect of a label on itself or on preceding labels. The parameters for the design featuring weak label interdependencies are
\begin{equation}
\label{eq:dgp_weak}
    \mathbf{b}^{(\text{w})} =  \left[\begin{array}{@{}r@{}}
        1.0 \\
        -2.5 \\
        -0.5
    \end{array}\right]
    \quad
    \mathbf{W}^{(\text{w})} = \left[\begin{array}{@{}rrr@{}}
        2.0 & 0.0 & 0.0 \\
        2.0 & 0.0 & 0.0 \\
        -3.0 & 0.0 & 0.0
    \end{array}\right]
    \quad
    \mathbf{C}^{(\text{w})} = \left[\begin{array}{@{}rrr@{}}
        \text{-}\phantom{0} & \text{-}\phantom{0} & \text{-}\phantom{0} \\
        1.0 & \text{-}\phantom{0} & \text{-}\phantom{0} \\
        2.5 & -3.0 & \text{-}\phantom{0}
    \end{array}\right].
\end{equation}
Important factors in determining whether a data generating process exhibits strong or weak label interdependencies are the magnitude of the elements of \ma{C} and the interplay of the signs of the effects in \ma{W} and \ma{C}. To illustrate this point, the label probabilities are visualized for both designs in Figure~\ref{fig:weak_strong}. The right panel shows that, unlike the probabilities associated with strong interdependencies, the probabilities in the data generating process with weak interdependencies exhibit a monotonic relationship with $x_1$. In Section~\ref{sec:conditionaldeps}, we discuss approaches for detecting strong label interdependencies.

{\bf Reversed label order.}
This design uses a data-generating process akin to the one with strong label interdependencies as described in \eqref{eq:dgp_strong}. The modified design features 200 observations and 6 labels. The exact parameters are detailed in Appendix~\ref{app:sim_designs}. To assess the impact of an incorrect label order on the classifier chain network, the label order is reversed after generating the data.

{\bf Sequential label realization.}
One of the aspects that differentiates the classifier chain network from the classifier chain \citep{read2009ecc, read2011ecc} is the assumption regarding label dependencies. In the classifier chain network, labels are assumed to depend on each other through their probabilistic distributions, rather than their observed values. Using the parameters from the reversed label order design, this data generating process adopts a sequential approach, where the label probabilities are transformed into binary outcomes before computing their effect on subsequent label probabilities.

{\bf Increased label count.}
The number of parameters that require estimation in the classifier chain network is $L(m + 1) + L(L-1)/2$. In the weak and strong simulation designs, the ratio of observations to estimated parameters is favorable at 13.3 ($200/15$). In the reversed label order and sequential label realization designs, this ratio reduces to 5.1 ($200/39$). To evaluate performance under a less favorable ratio, this design combines the 3-label design with strong interdependencies the and 6-label design into a 9-label design, reducing the ratio to 2.8 ($200/72$).

\subsection{Methods and Evaluation}
\label{subsec:sim_methods}
To evaluate the performance of the classifier chain network, we compare it  against a diverse set of multi-label classification methods, including both parametric and nonparametric methods. Our study primarily focuses on comparing methods that can be used individually or as components within ensembles, rather than on the ensemble methods themselves, such as the ensemble classifier chain \citep{read2009ecc, read2011ecc} and random forests \citep{breiman2001random}. By doing so, we aim to better understand the standalone capabilities of these methods and their potential contributions to more complex models. Each method is evaluated using multiple metrics to capture variations in performance across different measures. Section~\ref{subsubsec:sim_methods} details the specific settings of the classifier chain network and provides descriptions of the alternative methods and their tuning parameters, while Section~\ref{subsubsec:evaluation} covers the evaluation methodology.

\subsubsection{Methods}
\label{subsubsec:sim_methods}
We construct the classifier chain network according to the modeling choices outlined in Section~\ref{sec:ccn_choices}, using the binary cross-entropy loss function for the individual label predictions. For the initialization of the input parameters for Algorithm~\ref{alg:ccn_BFGS}, we use 10 random starts in addition to the informed initialization, as described in Section~\ref{sec:opt_tuning}. The tuning parameters are selected using a grid search with cross-validation, with candidate values for $q$ being $\{ 1, 1.5, 2, 3, 5 \}$ and for $\lambda$ being $\{ 0.0001, 0.001, 0.01, 0.05, 0.1, 0.25 \}$. Below, we list the selected benchmark methods.

{\bf Binary relevance (BR) and classifier chain (CC).}
To match the structure of the classifier chain network, these methods use a penalized logistic regression as the classifier. Candidate values for the regularization parameter $\lambda$ are $\{ 0.0001, 0.001, 0.01, 0.05, 0.1, 0.25 \}$. Both methods are part of the \emph{scikit-learn} package \citep{scikit-learn}.

{\bf AdaBoost.MH (Ada).}
\citet{schapire1998adaboost} demonstrate that minimizing the Hamming loss in multi-label classification using boosting is equivalent to a binary relevance approach, where a separate boosting model is trained for each label $\ell$. The resulting algorithm, AdaBoost.MH, has proven particularly effective in multi-label classification tasks. For instance, \citet{bogatinovski2022comprehensive} rank it among the top five performing methods, and notably, it is the only one of these that is not a parallel ensemble of independent multi-label classifiers. The base classifier is a decision stump and the tuning parameter is the number of consecutive stumps, with candidate values $\{ 25, 50, 75, 100, 125 \}$. We implemented this method using the \emph{scikit-learn} package.

{\bf Multi-label $k$-nearest neighbors (ML$k$NN).} 
This nonparametric method extends the classical $k$-nearest neighbors algorithm to multi-label classification \citep{zhang2005ml-knn}. Instead of predicting a single label, it assigns multiple labels to an observation based on the labels of its $k$ nearest neighbors, effectively capturing the relationship between labels in a local neighborhood. In this study, candidate values for the number of nearest neighbors range from 5 to $\lceil n^{1/2} \rceil=15$. The implementation is part of the \emph{scikit-multilearn} package \citep{szymański2018scikitmlc}.

{\bf Multi-label twin support vector machine (MLTSVM).}
Introduced by \citet{chen2016mltsvm}, this method extends twin support vector machines \citep{jayadeva2007twinsvm} by using multiple nonparallel hyperplanes to model label interdependencies in multi-label classification tasks. The intersections of these hyperplanes indicate regions where observations have multiple labels. The associated loss function balances model complexity and classification accuracy, with the penalty parameter $c_k \in \{0.75, 1, 1.25, 1.5, 1.75\}$ determining the weight of the classification error and the regularization parameter $\lambda \in \{0, 0.5, 1, 1.5, 2\}$ controlling overfitting. The implementation is part of the \emph{scikit-multilearn} package.

{\bf Random $k$-labelsets (RA$k$EL).}
The label powerset approach to multi-label classification \citep{boutell2004learning} considers each combination of labels as a unique class, allowing the use of a multi-class classifier. However, as the number of labels $L$ increases, the number of unique combinations grows exponentially, making this approach computationally infeasible. To address this, \citet{tsoumakas2011random} introduced the random $k$-labelsets algorithm, which creates overlapping subsets of the original labels and fits a label powerset classifier to each subset. In our simulation study, we use a penalized multinomial logistic regression as the classifier, with candidate values for the regularization parameter $\lambda$ being $\{ 0.0001, 0.001, 0.01, 0.05, 0.1, 0.25 \}$. The labelset size is set to 2 for $L=3$ and 3 for $L>3$, and the number of models is set to $2L$. The implementation is part of the \emph{scikit-multilearn} package.

\subsubsection{Evaluation}
\label{subsubsec:evaluation}
To assess the performance of the methods discussed, we use the metrics outlined in Section~\ref{sec:opt_tuning}. We use multiple metrics to capture potential differences between methods, as each may optimize different performance aspects \citep{dembczynski2010regret}. Probabilistic predictions are available only for the classifier chain network, binary relevance, classifier chain, and AdaBoost.MH; thus, negative log-likelihood is only reported for these methods.

The tuning parameters are determined through 5-fold cross-validates grid searches, with the scoring metric matching the evaluation metric. Out-of-sample performance is assessed on a validation set of 1,000 observations. The simulations are repeated 200 times to obtain performance statistics. Additionally, methods requiring a specific label order are provided with the correct order, unless stated otherwise in the simulation design (see Section~\ref{subsec:sim_designs}).

\subsection{Results}
\label{subsec:sim_results}
In Section~\ref{subsubsec:res_designs}, we discuss the results obtained for the different simulation designs presented in Section~\ref{subsec:sim_designs}. As not all methods produce probabilistic predictions, the results for the negative log-likelihood are discussed separately in Section~\ref{subsubsec:res_nll}. Section~\ref{subsubsec:res_conclusions} summarizes the findings of the simulation study.

\subsubsection{All Simulation Designs}
\label{subsubsec:res_designs}
\begin{figure}[t]
\includegraphics{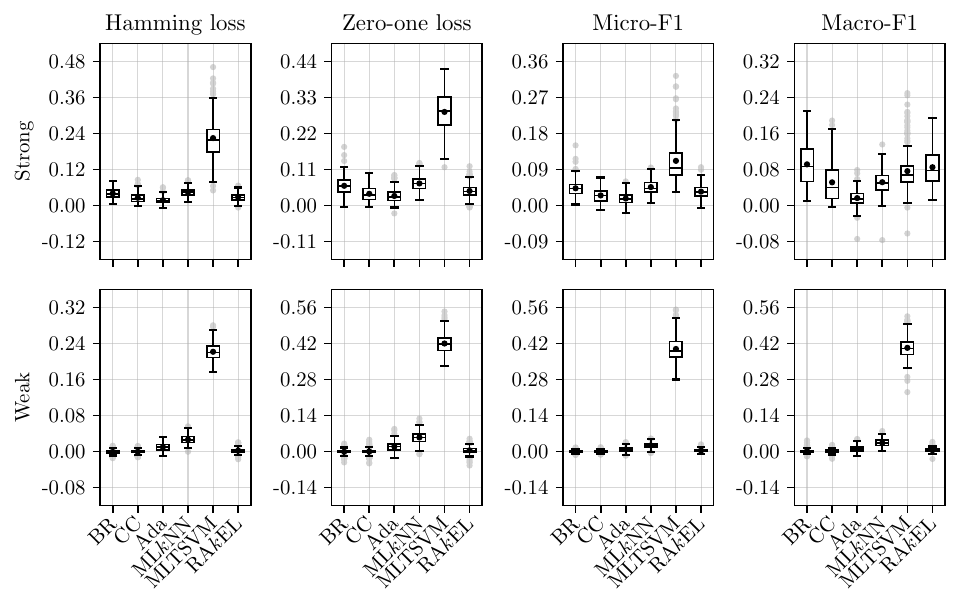}
\centering
\caption{Boxplots of the difference in the performance between the classifier chain network and the benchmark methods discussed in Section~\ref{subsec:sim_methods}. The differences are computed such that a positive value indicates improved performance of the classifier chain network over the corresponding method. The two simulation designs are the strong (top) and weak (bottom) label interdependencies. \label{fig:sim_res_a}}
\end{figure}

\begin{figure}[t]
\includegraphics{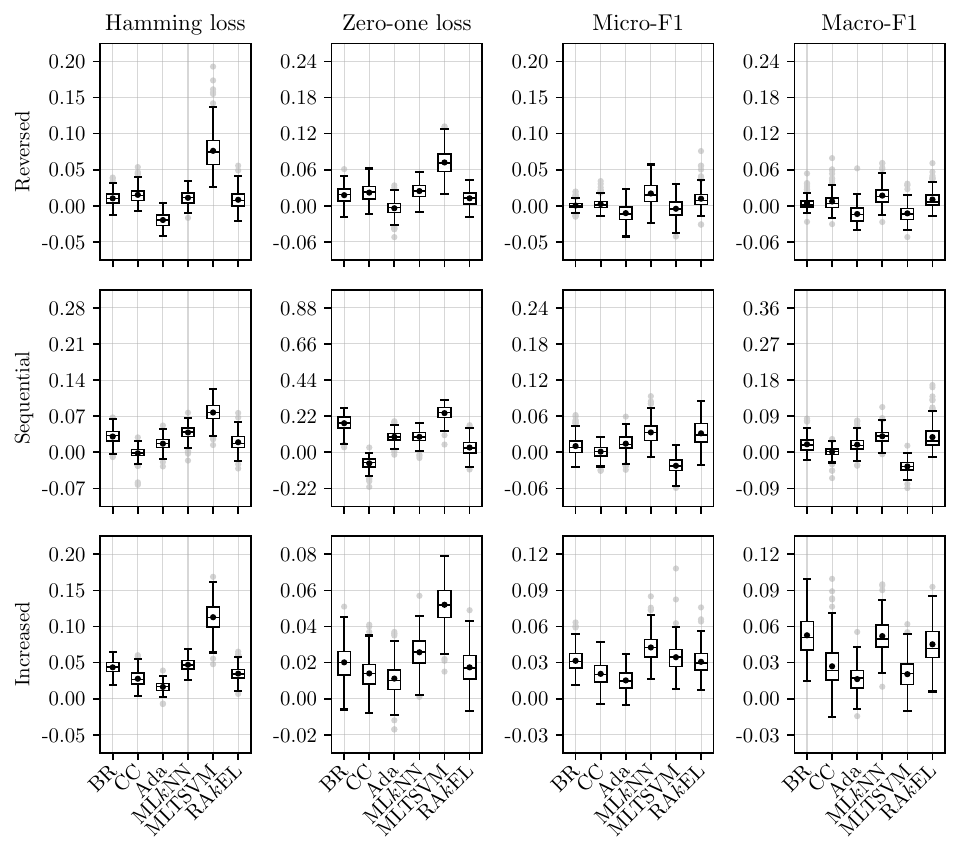}
\centering
\caption{The results for the simulation designs with reversed label order, sequential label realization, and increased label count. See Figure~\ref{fig:sim_res_a} for explanatory notes.\label{fig:sim_res_b}}
\end{figure}

We discuss the results for each simulation design in turn. Figures~\ref{fig:sim_res_a} and \ref{fig:sim_res_b} provide visualizations of the performance differences between the classifier chain network and the benchmark methods. For clarity, differences are presented so that a positive value indicates that the classifier chain network outperforms the benchmark. Table~\ref{tab:sim_res} summarizes the average scores of all methods for the performance metrics.

\begin{table}[p]
\centering
\sisetup{detect-weight,
         mode=text,
}
\begin{tabularx}{\linewidth}{
  c
  p{4.15cm}
  S[table-format=0.3, table-space-text-post = {*}, table-column-width=1.17cm]
  S[table-format=0.3, table-space-text-post = {*}, table-column-width=1.17cm]
  S[table-format=0.3, table-space-text-post = {*}, table-column-width=1.17cm]
  S[table-format=0.3, table-space-text-post = {*}, table-column-width=1.17cm]
  S[table-format=0.3, table-space-text-post = {*}, table-column-width=1.17cm]
  S[table-format=0.3, table-space-text-post = {*}, table-column-width=1.17cm]
  S[table-format=0.3, table-space-text-post = {*}, table-column-width=1.17cm]
}
\toprule
& Metric & CCN & BR & CC & \text{Ada} & ML$k$ & MLT & RA$k$ \\
\midrule
\multirow{5}*{\rotatebox{90}{Strong}}
& Hamming loss $\downarrow$ & \B 0.251 & 0.291$^{*}$ & 0.277$^{*}$ & 0.268$^{*}$ & 0.295$^{*}$ & 0.475$^{*}$ & 0.279$^{*}$ \\
& Zero-one loss $\downarrow$ & \B 0.570 & 0.631$^{*}$ & 0.606$^{*}$ & 0.600$^{*}$ & 0.638$^{*}$ & 0.856$^{*}$ & 0.615$^{*}$ \\
& Neg. log-likelihood $\downarrow$ & \B 0.514 & 0.559$^{*}$ & 0.544$^{*}$ & 0.580$^{*}$ \\
& Micro-F1 $\uparrow$ & \B 0.756 & 0.713$^{*}$ & 0.730$^{*}$ & 0.738$^{*}$ & 0.710$^{*}$ & 0.644$^{*}$ & 0.721$^{*}$ \\
& Macro-F1 $\uparrow$ & \B 0.723 & 0.632$^{*}$ & 0.672$^{*}$ & 0.707$^{*}$ & 0.671$^{*}$ & 0.647$^{*}$ & 0.638$^{*}$ \\
\midrule
\multirow{5}*{\rotatebox{90}{Weak}}
& Hamming loss & 0.145 & \B 0.144$^{\dagger}$ & \B 0.144 & 0.155$^{*}$ & 0.171$^{*}$ & 0.366$^{*}$ & 0.147$^{*}$ \\
& Zero-one loss & \B 0.366 & \B 0.366 & \B 0.366 & 0.386$^{*}$ & 0.421$^{*}$ & 0.786$^{*}$ & 0.371$^{*}$ \\
& Neg. log-likelihood & \B 0.323 & \B 0.323$^{*}$ & \B 0.323$^{*}$ & 0.468$^{*}$ \\
& Micro-F1 & 0.853 & \B 0.854 & 0.853 & 0.845$^{*}$ & 0.829$^{*}$ & 0.454$^{*}$ & 0.849$^{*}$ \\
& Macro-F1 & \B 0.840 & 0.838$^{*}$ & 0.838$^{*}$ & 0.829$^{*}$ & 0.806$^{*}$ & 0.437$^{*}$ & 0.834$^{*}$ \\
\midrule
\multirow{5}*{\rotatebox{90}{Reversed}}
& Hamming loss & 0.329 & 0.340$^{*}$ & 0.345$^{*}$ & \B 0.310$^{\dagger}$ & 0.340$^{*}$ & 0.405$^{*}$ & 0.338$^{*}$ \\
& Zero-one loss & 0.874 & 0.892$^{*}$ & 0.896$^{*}$ & \B 0.870$^{\dagger}$ & 0.899$^{*}$ & 0.947$^{*}$ & 0.887$^{*}$ \\
& Neg. log-likelihood & \B 0.577 & 0.587$^{*}$ & 0.593$^{*}$ & 0.597$^{*}$ \\
& Micro-F1 & 0.711 & 0.710 & 0.708$^{*}$ & \B 0.721$^{\dagger}$ & 0.694$^{*}$ & 0.715$^{\dagger}$ & 0.701$^{*}$ \\
& Macro-F1 & 0.698 & 0.693$^{*}$ & 0.691$^{*}$ & \B 0.711$^{\dagger}$ & 0.681$^{*}$ & 0.710$^{\dagger}$ & 0.687$^{*}$ \\
\midrule
\multirow{5}*{\rotatebox{90}{Sequential}}
& Hamming loss & 0.347 & 0.377$^{*}$ & \B 0.345 & 0.363$^{*}$ & 0.385$^{*}$ & 0.424$^{*}$ & 0.366$^{*}$ \\
& Zero-one loss & 0.745 & 0.922$^{*}$ & \B 0.677$^{\dagger}$ & 0.839$^{*}$ & 0.839$^{*}$ & 0.983$^{*}$ & 0.773$^{*}$ \\
& Neg. log-likelihood & \B 0.597 & 0.618$^{*}$ & 0.629$^{*}$ & 0.628$^{*}$ \\
& Micro-F1 & 0.683 & 0.672$^{*}$ & 0.682 & 0.668$^{*}$ & 0.650$^{*}$ & \B 0.705$^{\dagger}$ & 0.651$^{*}$ \\
& Macro-F1 & 0.668 & 0.649$^{*}$ & 0.668 & 0.649$^{*}$ & 0.628$^{*}$ & \B 0.704$^{\dagger}$ & 0.631$^{*}$ \\
\midrule
\multirow{5}*{\rotatebox{90}{Increased}}
& Hamming loss & \B 0.279 & 0.323$^{*}$ & 0.307$^{*}$ & 0.296$^{*}$ & 0.326$^{*}$ & 0.392$^{*}$ & 0.314$^{*}$ \\
& Zero-one loss & \B 0.934 & 0.954$^{*}$ & 0.948$^{*}$ & 0.945$^{*}$ & 0.960$^{*}$ & 0.986$^{*}$ & 0.951$^{*}$ \\
& Neg. log-likelihood & \B 0.536 & 0.577$^{*}$ & 0.569$^{*}$ & 0.591$^{*}$ \\
& Micro-F1 & \B 0.743 & 0.712$^{*}$ & 0.723$^{*}$ & 0.728$^{*}$ & 0.701$^{*}$ & 0.709$^{*}$ & 0.713$^{*}$ \\
& Macro-F1 & \B 0.727 & 0.674$^{*}$ & 0.700$^{*}$ & 0.711$^{*}$ & 0.675$^{*}$ & 0.707$^{*}$ & 0.682$^{*}$ \\
\bottomrule\noalign{\smallskip}
\end{tabularx}
\caption{Out-of-sample performance for the methods discussed in Section~\ref{subsec:sim_methods} for each of the data generating processes described in Section~\ref{subsec:sim_designs}. Additional abbreviations are ML$k$ (ML$k$NN), MLT (MLTSVM), and RA$k$ (RA$k$EL). Performance measures for which lower is better are the hamming loss, zero-one loss, and negative log-likelihood. Those for which higher is better are the micro-F1 and macro-F1 scores. The symbol $\dagger$ ($*$) indicates a statistically significant improvement (degredation) over the classifier chain network (Wilcoxon signed-rank test at a 5\% significance level).\label{tab:sim_res}}
\end{table}

{\bf Baseline designs.} 
In the design with strong label interdependencies, the classifier chain network consistently outperforms the other methods across all performance metrics (see the first row of Figure~\ref{fig:sim_res_a}). However, for weak label interdependencies, the results differ substantially: the classifier chain network, binary relevance, classifier chain, and random $k$-labelsets show nearly identical performance (see the second row of Figure~\ref{fig:sim_res_a}). The weak conditional dependencies enable binary relevance to rank among the top performers, which is in line with the findings of \citet{luaces2012binary}. Additionally, while the performance difference is almost negligible, binary relevance outperforms the classifier chain network in a statistically significant number of cases (see the second panel of Table~\ref{tab:sim_res}). A possible explanation for this could be reduced estimation variability resulting from having fewer estimated parameters.

{\bf Reversed label order.}
The methods relying on the ordering of the labels are the classifier chain network and the classifier chain. Despite using a reversed label order, the classifier chain network ranks second in terms of the Hamming and zero-one losses, and third for both micro-F1 and macro-F1 scores (see the first row of Figure~\ref{fig:sim_res_b} and the third panel of Table~\ref{tab:sim_res}). Furthermore, the classifier chain network consistently outperforms the classifier chain, highlighting the advantage of jointly estimating model parameters even under suboptimal conditions. It is important to note that a reversed label order represents the most severe form of order misspecification and is unlikely to occur with a larger number of labels. For example, with six labels, the probability of a completely reversed label order is only 0.14\% when randomized.

{\bf Sequential label realization.}
In this scenario, the classifier chain network and the classifier chain show similar performance in terms of the Hamming loss (see the second row of Figure~\ref{fig:sim_res_b} and the fourth panel of Table~\ref{tab:sim_res}). Although the data-generating process aligns with the modeling assumptions of the classifier chain, it only outperforms the classifier chain network in terms of zero-one loss in a statistically significant number of simulations.  Consistent with the reversed label order results, the multi-label twin support vector machine performs best in terms of micro-F1 and macro-F1 scores, while performing relatively poorly on Hamming and zero-one losses. Conversely, the performance of AdaBoost.MH is negatively impacted by the change in the data generating process.

{\bf Increased label count.}
The increased number of labels does not change the performance gap between the classifier chain network and most other methods (see the third row of Figure~\ref{fig:sim_res_b} and the fifth panel of Table~\ref{tab:sim_res}). Consistent with the baseline simulation with strong label interdependencies, the classifier chain network outperforms the other methods across all performance metrics. However, unlike in the baseline simulation, the multi-label twin support vector machine has narrowed the performance gap, demonstrating greater competitiveness in scenarios with a higher label count.

\subsubsection{Negative Log-Likelihood}
\label{subsubsec:res_nll}

\begin{figure}[t]
\includegraphics{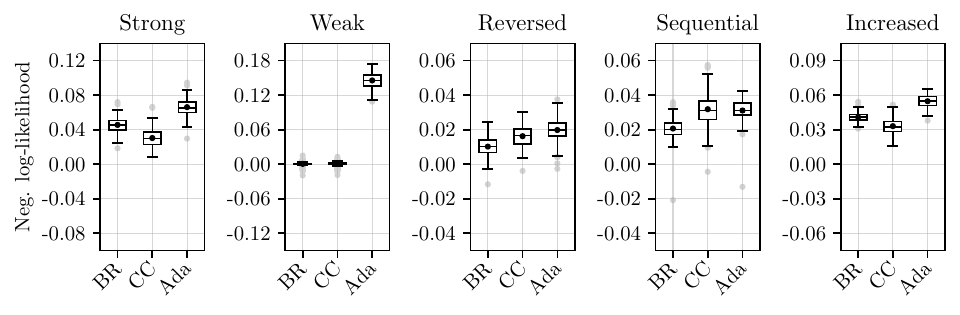}
\centering
\caption{The results for the negative log-likelihood for all simulation designs. See Figure~\ref{fig:sim_res_a} for explanatory notes.\label{fig:sim_res_nll}}
\end{figure}

Figure~\ref{fig:sim_res_nll} shows the differences in out-of-sample negative log-likelihood between the classifier chain network and binary relevance, the classifier chain, and AdaBoost.MH. Across all simulation designs, the classifier chain network consistently outperforms the other methods (see Table~\ref{tab:sim_res}). Unlike the other performance metrics, which measure prediction correctness, the negative log-likelihood assesses the certainty of each prediction. The modeling choices for the classifier chain network highlighted in Section~\ref{sec:ccn_choices} facilitate direct optimization of this metric, especially when a small value for $q$ is selected. Indeed, the simulation results reveal that values of $q \in \{1, 1.5\}$ were most frequently chosen during cross-validation.

\subsubsection{Conclusions from the Simulations}
\label{subsubsec:res_conclusions}
The classifier chain network demonstrates flexibility and exceptional performance across a range of simulation designs and performance metrics. Notably, it consistently matches or exceeds the benchmark methods in terms of the Hamming loss, even when the modeling assumptions do not align with the data generating process. The only exception is when the label order is reversed. That is, when there is severe label order misspecification. The results for the negative log-likelihood further underscore the importance of matching the loss function with the desired performance metric. Overall, the findings presented in this section make a compelling case for the use of the classifier chain network for multi-label classification.

\section{Detecting Conditional Dependencies}
\label{sec:conditionaldeps}
The simulation results presented in Section~\ref{subsubsec:res_designs} highlight scenarios where binary relevance outperforms the alternative methods considered. This difference in performance is influenced by the nature of the label interdependencies. Therefore, it is important to evaluate the suitability of a multi-label classification method before applying it to a given problem. For instance, an incorrect label order in a classifier chain can lead to the erroneous conclusion that strong label interdependencies are absent. This is illustrated by the negligible difference between the classifier chain network and binary relevance observed for the micro-F1 and macro-F1 scores in the first row of Figure~\ref{fig:sim_res_b}.

To further explore this topic, we consider conditional independency. That is, labels $\ell$ and $k$ are considered conditionally independent if the probability of observing label $\ell$, given label $k$, is equal to the unconditional probability, that is,
\begin{equation}
\label{eq:cond_indep}
    P(y_{i\ell}=1| y_{ik}, \ma{x}_i) = P(y_{i\ell}=1| \ma{x}_i),
\end{equation}
as noted by \citet{dembczynski2012label}. When this condition holds for all pairs of labels, binary relevance becomes a more efficient choice, as it typically requires fewer parameters than other multi-label classification methods. Moreover, even if this condition does not hold, binary relevance may still be the most suitable method, as illustrated by the results for the data generating process with weak label interdependencies (see the second row of Figure~\ref{fig:sim_res_a} and second panel of Table~\ref{tab:sim_res}).

Existing literature recognizes the importance of detecting strong conditional dependencies in advance \citep[e.g.,][]{dembczynski2012label, read2021classifier}. However, available methods to characterize the strength of the label interdependencies within a data set often do not include the explanatory variables $\ma{x}_i$ in \eqref{eq:cond_indep}, thus ignoring these variables as a potential source of correlation. In Section~\ref{subsec:label_dep_measures}, we review three existing approaches for quantifying label interdependencies and propose a new approach to address their limitations. To evaluate the effectiveness of these measures, we conduct a simulation study in Section~\ref{subsec:label_dep_measures_sim}.

\subsection{Measures}
\label{subsec:label_dep_measures}
The first measure we discuss, label density \citep{tsoumakas2010mining}, is defined as the average fraction of positive labels per observation, that is,
\begin{equation*}
    \textrm{Label density} = \frac{1}{nL} \sum_{i=1}^n \sum_{\ell=1}^L y_{i\ell}.
\end{equation*}
The second measure, label dependency \citep{luaces2012binary}, is a weighted average of the correlations between all pairs of labels, using the co-occurrences of the respective labels as weights. It is computed as
\begin{equation*}
    \textrm{Label dependency} = \frac{\sum_{\ell<k} |\rho_{k\ell}| \sum_{i=1}^n y_{ik} y_{i\ell}}{\sum_{\ell<k} \sum_{i=1}^n y_{ik} y_{i\ell}},
\end{equation*}
where $\rho_{k\ell}$ denotes the correlation coefficient between labels $k$ and $\ell$. These measures share a common limitation: they are not invariant to label negation. That is, if we reverse the arbitrary coding of labels, for instance, ``purchase'' into ``no purchase'', the measures yield different values. 

The third measure, the fraction of unconditionally dependent label pairs at 99\% confidence \citep{tenenboim2010identlabeldep, moyano2018review}, does not suffer from the label negation limitation. It uses a Chi-square test for independence of two binary variables and is computed as
\begin{equation*}
    \textrm{Unconditional dependency} = \frac{1}{L(L-1)/2} \sum_{k<\ell} \indicator{\textrm{$p$-value}_{k\ell} \leq 0.01}.
\end{equation*}
Note that none of these three measures condition on the variables in $\ma{X}$, even though these may cause nonzero correlations between labels. Therefore, we propose a different measure that does take into account the variables in \ma{X}.

To detect conditional dependencies between labels we consider the estimated difference in out-of-sample performance between two classifiers. The first classifier uses only the information in $\ma{x}_i$ to predict $y_{i\ell}$, while the second uses both $\ma{x}_i$ and the other labels $\{ y_{ik} : 1 \leq k \leq L, k \neq \ell \}$. These classifiers are denoted by $\mathcal{M}_\ell(\ma{X})$ and $\mathcal{M}_\ell(\ma{Y}_{-\ell}, \ma{X})$, respectively. If label $\ell$ is conditionally independent of the other labels given the variables in \ma{X}, the performance difference between the two classifiers should be minimal. Conversely, an improvement in the relative performance of $\mathcal{M}_\ell(\ma{Y}_{-\ell}, \ma{X})$ indicates the presence of conditional dependencies between label $\ell$ and one or more of the remaining labels. The proposed measure for conditional dependency within the data is
\begin{equation}
\label{eq:cond_dep_score}
    \textrm{Performance}( \{ \mathcal{M}_\ell(\ma{Y}_{-\ell}, \ma{X}) : 1 \leq \ell \leq L \} ) - \textrm{Performance} ( \{ \mathcal{M}_\ell(\ma{X}) : 1 \leq \ell \leq L \} ),
\end{equation}
where the performance can be measured using any of the multi-label classification evaluation metrics presented in Section~\ref{sec:opt_tuning}. The difference can be estimated via (nested) cross-validation. Unlike multi-label classification methods, this approach does not require modeling the outcomes for the other labels. Instead, it exploits their true values. Using the information provided by the other labels avoids complications such as specifying a label order for classifier chains and enables the detection of conditional dependencies under optimal conditions.

\subsection{Simulation Study}
\label{subsec:label_dep_measures_sim}
To evaluate the effectiveness of the different measures discussed, we conduct a simulation study with 100 different data generating processes. The aim of these simulations is to determine which methods can identify scenarios where the classifier chain network provides an advantage over binary relevance in terms of the Hamming loss. We focus on the Hamming loss because, in the simulation design with weak label interdependencies, binary relevance outperformed the classifier chain network in a statistically significant number of cases (see the second panel of Table~\ref{tab:sim_res}). For our proposed method to measure conditional dependency, we model each label using penalized logistic regression. Out-of-sample performance is estimated using 10-fold cross-validation and the regularization parameter is selected using 5-fold cross-validation.

\begin{figure}[t]
\includegraphics{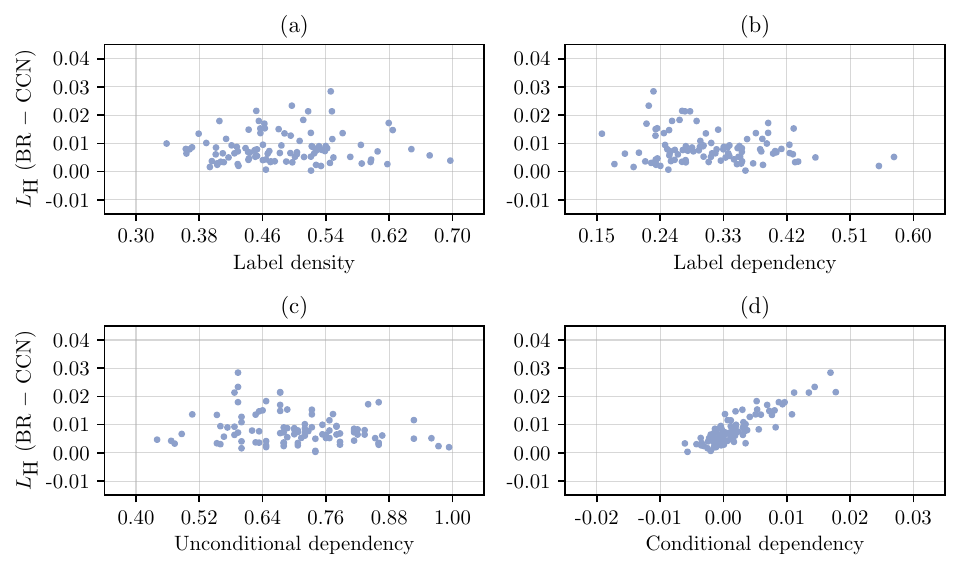}
\centering
\caption{Scatter plots of the excess Hamming loss of binary relevance over the classifier chain network and four different scores to quantify the strength of the interdependencies between the labels. The Spearman rank-order correlation coefficients and their $p$-values are: (a) $0.03$ ($p = 0.761$); (b) $-0.07$ ($p = 0.491$); (c) $-0.15$ ($p = 0.136$); and (d) $0.79$ ($p<0.001$).\label{fig:label_dep_scores}}
\end{figure}

The parameter settings for the data generating processes are based on the six-label design discussed in Section~\ref{subsec:sim_designs} and defined in Appendix~\ref{app:sim_designs}. To create data sets with varying levels of conditional dependency, each parameter in \ma{b}, \ma{W}, and \ma{C} is drawn from a normal distribution with a mean of zero and a standard deviation of four. The resulting parameters are validated by ensuring that none of the labels are excessively imbalanced, with no majority class comprising more than 85\% of the observations, on average. Each data generating process is used to generate 10 training data sets, each containing 200 observations. To evaluate the difference in performance between the classifier chain network and binary relevance, validation sets of 1,000 observations are created. For each data generating process, we calculate the average difference in Hamming loss between binary relevance and the classifier chain network. Additionally, we compute the average values for the label density, label dependency, unconditional dependency, and conditional dependency measures using the training data. The results of these simulations are presented in Figure~\ref{fig:label_dep_scores}, where a useful measure should be strongly correlated with the excess performance.

The results indicate that label density is not significantly correlated with the excess performance of the classifier chain network. This is expected, as negating \ma{Y} to $\neg\ma{Y}$ changes the label density to one minus the original value, while leaving the Hamming loss unchanged. A similar issue arises with label dependency. The unconditional dependency measure detects correlations but, because it does not condition on the variables in \ma{X}, it shows no significant correlation with the classifier chain network's performance relative to binary relevance. In contrast, our proposed method for detecting conditional dependency demonstrates a strong relationship. With a Spearman rank-order correlation coefficient of 0.79, it is the most reliable indicator of when methods that explicitly model label interdependencies are advantageous.

\section{Application}
\label{sec:application}
To illustrate how the classifier chain network works in practice, we used the \emph{emotions} data set from \cite{trohidis2008multilabel}. This data set contains emotional responses recorded by three experts while listening to 593 sound clips from various musical genres. Each clip is tagged with one or more of six emotions: \emph{quiet-still}, \emph{sad-lonely}, \emph{angry-aggressive}, \emph{amazed-surprised}, \emph{happy-pleased}, and \emph{relaxing-calm}. Additionally, the data set includes 72 explanatory variables, consisting of 8 rhythmic features and 64 timbre features. Further details can be found in the original paper.

To prepare the data for analysis, we standardized the explanatory variables and applied principal component analysis, retaining the first 29 components to capture over 90\% of the explained variance. The conditional dependency score in \eqref{eq:cond_dep_score} was calculated as 0.065, considerably higher than zero, indicating that label correlations exist which are not fully explained by the explanatory variables alone.

\begin{figure}[t]
\includegraphics{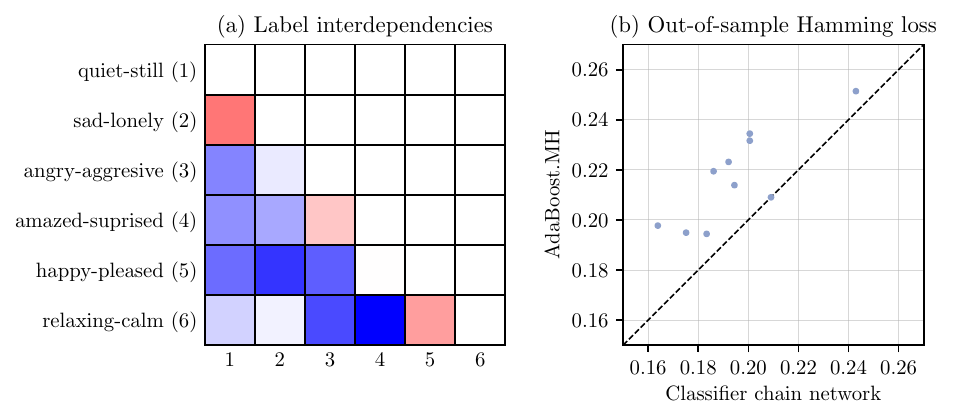}
\centering
\caption{Emotions application. On the left, the estimated label interdependencies, color-coded based on the values in $\widehat{\ma{C}}$, where blue (red) indicates a negative (positive) value. On the right, the out-of-sample Hamming losses across the ten testing samples (dots) for the classifier chain network versus AdaBoost.MH. \label{fig:application_results}}
\end{figure}

We determined the label order for the classifier chain network using the method proposed by \cite{jun2019conditional}, which relies on the conditional entropy of the labels. Specifically, labels were ordered by their conditional entropy, with those having lower conditional entropy placed at the beginning of the chain, in accordance with the third strategy proposed by the authors. The optimal tuning parameters were selected through 5-fold cross-validation, using the same candidate values as in our simulation study (see Section~\ref{subsubsec:sim_methods}), with the Hamming loss as evaluation metric. In the left panel of Figure~\ref{fig:application_results}, we visualize the estimated label interdependencies derived from the classifier chain network. Each cell is color-coded based on its value in $\widehat{\ma{C}}$, with blue representing negative values and red positive ones. For instance, the labels \emph{quiet-still} and \emph{relaxing-calm} are positively correlated ($\rho = 0.30$), yet, after controlling for explanatory variables, the estimated direct effect of \emph{quiet-still} on \emph{relaxing-calm} is weakly negative. This suggests that the positive correlation between these emotions can be attributed largely to measurable features of the sound clips.

We also compared the performance of the classifier chain network with AdaBoost.MH, using the Hamming loss as the evaluation metric. AdaBoost.MH was selected due to its direct optimization of the Hamming loss \citep{schapire1998adaboost} and its superior performance in our simulations when label order was misspecified (see Table~\ref{tab:sim_res}). Performance was evaluated via 10-fold cross-validation. In the right panel of Figure~\ref{fig:application_results}, we display the out-of-sample Hamming-losses, with the classifier chain network on the horizontal axis and AdaBoost.MH on the vertical axis. The classifier chain network outperformed AdaBoost.MH in the majority of test samples, achieving lower values for the Hamming loss overall.

\section{Conclusion}
\label{sec:conclusion}
In this study, we introduced the classifier chain network, a generalization of the traditional classifier chain \citep{read2009ecc, read2011ecc} for multi-label classification. This method integrates the chain of binary classifiers into a single network, allowing for joint estimation of parameters that account for the influence of earlier predictions on subsequent labels. This structure provides a more coherent modeling of label interdependencies.

We conducted a comprehensive simulation study comparing the classifier chain network to a range of multi-label classification approaches, including both parametric and nonparametric methods. The results demonstrated a clear advantage of the classifier chain network across diverse scenarios, while also highlighting the importance of selecting models that align with the evaluation metrics relevant to specific applications. Additionally, we evaluated various descriptive statistics aimed at quantifying the strength of the label interdependencies within a data set. We proposed a new conditional dependency measure and appraised its performance in a second simulation study. Our measure showed a significant improvement in detecting potential label interdependencies.

In applying the classifier chain network to empirical data, we further demonstrated the interpretability of its model parameters, underscoring its potential as a useful tool for both prediction and insight generation in multi-label classification tasks. This positions the classifier chain network as a strong alternative to existing methods, particularly for practitioners seeking interpretable models that capture complex label relationships.

Future research could explore different specifications of the relationships between labels and explanatory variables and between labels. For example, allowing each label to depend on distinct sets of explanatory variables (as suggested by \citealp{weng2020label}) increases flexibility. Furthermore, although the current classifier chain network is not a deep learning method, incorporating hidden layers between output nodes could lead to a more general architecture, bearing similarities to the method proposed by \cite{cisse2016adios}. This would increase the method's capacity to capture complex patterns, though at the cost of higher complexity and potentially reduced interpretability. Finally, the classifier chain network can be used as a component of an ensemble, a strategy employed by \cite{read2009ecc, read2011ecc} to deal with the uncertainty in the label order.

\section*{Computational Details}
All experiments in this study were performed with Python 3.12. The optimization procedure for the classifier chain network was implemented in C\texttt{++} using the linear algebra library \emph{Eigen} \citep{eigen} and interfaced with Python using \emph{pybind11} \citep{pybind11}. A software package called \emph{CCNPy} implementing the classifier chain network is available at \url{https://github.com/djwtouw/CCNPy}. \emph{Replication files will be made publicly available upon publication and a link will be provided here.}

\FloatBarrier
\bibliographystyle{apalike}
\bibliography{references}

\appendix
\section{CCN Gradients}
\label{appendix:gradients}
This appendix details the derivation of the gradient required for the minimization of the loss function associated with the classifier chain network in \eqref{eq:ccn_loss}. For ease of exposition, the following notation is used. First, the gradient is considered on a per-observation basis, such that
\begin{equation*}
    \nabla L(\ma{b}, \ma{W}, \ma{C}) = \sum_{i=1}^n \nabla L_i(\ma{b}, \ma{W}, \ma{C})
\end{equation*}
and we omit the summation from the derivations. Second, the gradient is decomposed into the partial derivates with respect to the parameter groups $\ma{b}$ $\ma{W}$, and $\ma{C}$. Third, the following abbreviations are used
\begin{equation*}
    p_{i\ell} = \alpha(\theta_{i\ell}) \quad \text{and} \quad
    \theta_{i\ell} = b_\ell + \ma{x}_i^\top \ma{w}_\ell + \sum_{\ell'=1}^{\ell - 1} \dep{\ell}{k}.
\end{equation*}
The definitions of the classifier chain network and its loss function in Section~\ref{sec:ccn} are in general terms. 
However, the choice for the dependency structure in \eqref{eq:ccn_dependency} facilitates a more straightforward derivation of the gradient. Specifically,
\begin{equation*}
    \fracpartial{\theta_{i\ell}}{p_{ik}} = c_{\ell k} \quad \text{and} \quad \fracpartial{\theta_{i\ell}}{c_{\ell k}} = \fracpartial{\theta_{i\ell'}}{c_{\ell' k}} = p_{ik}.
\end{equation*}
Hence, these are used in the derivations that follow. The resulting partial derivatives are
\begin{align*}
    % b
    \fracpartial{L_i(\ma{b}, \ma{W}, \ma{C})}{b_L} &= 
    \fracpartial{L_i(\ma{b}, \ma{W}, \ma{C})}{p_{iL}} 
    \fracpartial{p_{iL}}{\theta_{iL}} \\
    \fracpartial{L_i(\ma{b}, \ma{W}, \ma{C})}{b_{L-1}} &= 
    \left( \fracpartial{L_i(\ma{b}, \ma{W}, \ma{C})}{p_{i,L-1}} + c_{L,L-1} \fracpartial{L_i(\ma{b}, \ma{W}, \ma{C})}{b_L} \right)
    \fracpartial{p_{i,L-1}}{\theta_{i,L-1}} \\
    \fracpartial{L_i(\ma{b}, \ma{W}, \ma{C})}{b_{L-2}} &= \left( \fracpartial{L_i(\ma{b}, \ma{W}, \ma{C})}{p_{i,L-2}} + c_{L,L-2} \fracpartial{L_i(\ma{b}, \ma{W}, \ma{C})}{b_L} + c_{L-1,L-2} \fracpartial{L_i(\ma{b}, \ma{W}, \ma{C})}{b_{L-1}} \right) \fracpartial{p_{i,L-2}}{\theta_{i,L-2}} \\
    \vdots\\
    \fracpartial{L_i(\ma{b}, \ma{W}, \ma{C})}{b_{L-\ell}} &= \left( \fracpartial{L_i(\ma{b}, \ma{W}, \ma{C})}{p_{i,L-\ell}} + \sum_{k=0}^{\ell-1} c_{L-k,L-\ell} \fracpartial{L_i(\ma{b}, \ma{W}, \ma{C})}{b_{L-k}} \right) \fracpartial{p_{i,L-\ell}}{\theta_{i,L-\ell}} \\
    % W
    \fracpartial{L_i(\ma{b}, \ma{W}, \ma{C})}{\ma{w}_L} &= 
    \fracpartial{L_i(\ma{b}, \ma{W}, \ma{C})}{p_{iL}} \fracpartial{p_{iL}}{\theta_{iL}} \ma{x}_i \\
    \fracpartial{L_i(\ma{b}, \ma{W}, \ma{C})}{\ma{w}_{L-1}} &= \left( \fracpartial{L_i(\ma{b}, \ma{W}, \ma{C})}{p_{i,L-1}} \ma{x}_i + c_{L,L-1} \fracpartial{L_i(\ma{b}, \ma{W}, \ma{C})}{\ma{w}_L} \right) \fracpartial{p_{i,L-1}}{\theta_{i,L-1}} \\
    \fracpartial{L_i(\ma{b}, \ma{W}, \ma{C})}{\ma{w}_{L-2}} &= \left( \fracpartial{L_i(\ma{b}, \ma{W}, \ma{C})}{p_{i,L-2}} \ma{x}_i + c_{L,L-2} \fracpartial{L_i(\ma{b}, \ma{W}, \ma{C})}{\ma{w}_L} + c_{L-1,L-2} \fracpartial{L_i(\ma{b}, \ma{W}, \ma{C})}{\ma{w}_{L-1}} \right) \fracpartial{p_{i,L-2}}{\theta_{i,L-2}} \\
    \vdots\\
    \fracpartial{L_i(\ma{b}, \ma{W}, \ma{C})}{\ma{w}_{L-\ell}} &= \left( \fracpartial{L_i}{p_{i,L-\ell}} \ma{x}_i + \sum_{k=0}^{\ell-1} c_{L-k,L-\ell} \fracpartial{L_i(\ma{b}, \ma{W}, \ma{C})}{\ma{w}_{L-k}} \right) \fracpartial{p_{i,L-\ell}}{\theta_{i,L-\ell}} \\
    % C
    \fracpartial{L_i(\ma{b}, \ma{W}, \ma{C})}{c_{L,\ell'}} &= 
    \fracpartial{L_i(\ma{b}, \ma{W}, \ma{C})}{p_{iL}} \fracpartial{p_{iL}}{\theta_{iL}} p_{\ell'} \\
    \fracpartial{L_i(\ma{b}, \ma{W}, \ma{C})}{c_{L-1,\ell'}} &= \left( \fracpartial{L_i(\ma{b}, \ma{W}, \ma{C})}{p_{i,L-1}} p_{\ell'} + c_{L,L-1} \fracpartial{L_i(\ma{b}, \ma{W}, \ma{C})}{c_{L, \ell'}} \right) \fracpartial{p_{i,L-1}}{\theta_{i,L-1}} \\
    \fracpartial{L_i(\ma{b}, \ma{W}, \ma{C})}{c_{L-2,\ell'}} &= \left( \fracpartial{L_i(\ma{b}, \ma{W}, \ma{C})}{p_{i,L-2}} p_{\ell'} + c_{L,L-2} \fracpartial{L_i(\ma{b}, \ma{W}, \ma{C})}{c_{L,\ell'}} + c_{L-1,L-2} \fracpartial{L_i(\ma{b}, \ma{W}, \ma{C})}{c_{L-1,\ell'}} \right) \fracpartial{p_{i,L-2}}{\theta_{i,L-2}} \\
    \vdots\\
    \fracpartial{L_i(\ma{b}, \ma{W}, \ma{C})}{c_{L-\ell,\ell'}} &= \left( \fracpartial{L_i(\ma{b}, \ma{W}, \ma{C})}{p_{i,L-\ell}} p_{\ell'} + \sum_{k=0}^{\ell-1} c_{L-k,L-\ell} \fracpartial{L_i(\ma{b}, \ma{W}, \ma{C})}{c_{L-k,\ell'}} \right) \fracpartial{p_{i,L-\ell}}{\theta_{i,L-\ell}}.
\end{align*}

\section{Additional Simulation Designs}
\label{app:sim_designs}
The parameters for the simulation design with $L=6$ labels discussed in Section~\ref{subsec:sim_designs} are
\begin{equation*}
    \mathbf{b} =  \left[\begin{array}{@{}r@{}}
        1.0 \\
        3.0 \\
        0.5 \\
        0.0 \\
        0.0 \\
        0.0  
    \end{array}\right]
    \quad
    \mathbf{W} = \left[\begin{array}{@{}rrr@{}}
        2.0  & 0.0 & 0.0\\
        1.0  & 0.0 & 0.0\\
        -0.5 & 0.0 & 0.0\\
        -1.0 & 0.0 & 0.0\\
        -3.0 & 0.0 & 0.0\\
        1.0  & 0.0 & 0.0
    \end{array}\right] 
    \quad
    \mathbf{C} = \left[\begin{array}{@{}rrrrrr@{}}
        \text{-}\phantom{0} & \text{-}\phantom{0} & \text{-}\phantom{0} & \text{-}\phantom{0} & \text{-}\phantom{0} & \text{-}\phantom{0}  \\
        -4.0 & \text{-}\phantom{0} & \text{-}\phantom{0} & \text{-}\phantom{0} & \text{-}\phantom{0} & \text{-}\phantom{0} \\
        -1.0 &  0.0 & \text{-}\phantom{0}  & \text{-}\phantom{0} & \text{-}\phantom{0} & \text{-}\phantom{0} \\
        4.0  & -2.0 & -2.0  & \text{-}\phantom{0} & \text{-}\phantom{0} & \text{-}\phantom{0} \\
        0.0  & -2.0 & -6.0  & 6.0 & \text{-}\phantom{0} & \text{-}\phantom{0} \\
        0.0  &  0.0 &  6.0  & 0.0 & -6.0 & \text{-}\phantom{0}
    \end{array}\right].
\end{equation*}

\end{document}